\documentclass[runningheads]{llncs}
\usepackage{graphicx}
\usepackage{amsmath,amssymb} 
\usepackage{color}
\usepackage{floatrow}
\usepackage{subfig}
\newfloatcommand{capbtabbox}{table}[][\FBwidth]

\usepackage{hyperref}

\begin{document}
\pagestyle{headings}
\mainmatter
\def\ECCV18SubNumber{8}  

\title{PathGAN: Visual Scanpath Prediction with Generative Adversarial Networks} 

\titlerunning{ECCV-18 EPIC Workshop submission ID \ECCV18SubNumber}
\authorrunning{ECCV-18 submission ID \ECCV18SubNumber}

\author{
Marc Assens\inst{1} 
\and
Xavier Giro-i-Nieto\inst{2}\orcidID{0000-0002-9935-5332}
\and
\\
Kevin McGuinness\inst{1}
\and
Noel E. O'Connor\inst{1}\orcidID{0000-0002-4033-9135}
}

\authorrunning{M. Assens et al.}
%
\institute{Dublin City University, Glasnevin, Whitehall, Dublin 9, Ireland \and
Universitat Politecnica de Catalunya, 08034 Barcelona, Catalonia/Spain \\
\email{kevin.mcguinness@insight-centre.org, xavier.giro@upc.edu}}


\maketitle

\begin{abstract}
We introduce PathGAN, a deep neural network for visual scanpath prediction trained on adversarial examples. A visual scanpath is defined as the sequence of fixation points over an image defined by a human observer with its gaze. PathGAN is composed of two parts, the generator and the discriminator. Both parts extract features from images using off-the-shelf networks, and train recurrent layers to generate or discriminate scanpaths accordingly. In scanpath prediction, the stochastic nature of the data makes it very difficult to generate realistic predictions using supervised learning strategies, but we adopt adversarial training as a suitable alternative. Our experiments prove how PathGAN improves the state of the art of visual scanpath prediction on the iSUN and Salient360! datasets.

\end{abstract}
\section{Introduction}
\label{sec:motivation}

When a human observer looks at an image, he spends most of his time looking at specific regions \cite{porter2007,amor2016persistence}. He starts directing his gaze at a specific point and explores the image creating a sequence of fixation points that covers the salient areas of the image. This process can be seen as a resource allocation problem; our visual system decides where to direct its attention, in which order, and how much time will be spent in each location given an image.

Visual saliency prediction is the field of computer vision that focuses on estimating the image regions that attract human attention. The understanding of this process can provide clues on human image understanding, and has applications in domains such as image and video compression, transmission, and rendering. In order to train and evaluate saliency prediction models, there exist scientific datasets containing fixation points generated by human observers when exploring an image without any specific task in mind. They are traditionally captured with eye-trackers \cite{wilming2017extensive}, mouse clicks \cite{jiang2015salicon}, and webcams \cite{cvpr2016_Khosla}.

These fixation points have an important characteristic: stochasticity \cite{assens2017saltinet}. Different human observers can produce very different fixation points. Thus, researchers in the field of saliency prediction have traditionally aggregated fixations of multiple observers to generate a consistent representation called saliency map \cite{itti1998model}. A saliency map is a single channel image obtained by convolving a Gaussian kernel with each fixation. The result is a gray-scale heatmap that represents the probability of each pixel in an image being fixated by a human, and it is usually used as a soft-attention guide for other computer vision tasks. 

Because fixations are aggregated over the temporal dimension,  the saliency map representation loses all the temporal information. Thus, information like \textit{the parts of an image that are being fixated first} is not retained. Recent studies have shown some of the limitations of saliency maps and have raised the need for a representation that is also temporally-aware \cite{bylinskii2016should}. In some situations saliency maps fail to represent the relative importance of the different parts of an image, giving more relevance to small regions with text where humans spend a long time reading. We believe that the regions where a human first fixates might be more relevant, therefore they should have more weight in a soft-attention representation.
Another argument that favors temporally-aware saliency representations is the recent explosion of Virtual Reality technologies. It has brought new challenges regarding the usage of omni directional images (360-degree images), and it seems that solutions will depend on the use of temporal information.

Thus, there is an increasing demand for temporally-aware saliency representations such as scanpaths, and algorithms that are capable of working with them. Scanpaths as a temporally-aware saliency representation have received recent attention \cite{cerf2008predicting,salient360} and different datasets are available today.

Previous work on scanpath prediction shows that there are difficulties when working with very stochastic data \cite{assens2017saltinet}. One of the problems that has been found is that supervised learning algorithms using the MSE loss do not perform well for this task because the final prediction tends to be the average of all the possible predictions \cite{mathieu2015deep}. When predicting scanpaths, the average prediction tends to be always in the center.
Recently, Goodfellow et al. \cite{goodfellow2014generative} proposed a framework to create generative functions via an adversarial process, in which two models are trained simultaneously: a generative model G that captures the data distribution, and a discriminative model D that estimates the probability that a sample comes from the training data rather than G. The training procedure for G is to maximize the probability of D making a mistake. This process allows models to generate realistic predictions even when the data has very complicated distributions. This framework seems a suitable technique for the generation of realistic scan paths. 

This paper explores an end-to-end solution for omni directional scanpath prediction using conditional adversarial training. We show that this framework is suitable for this task and it significantly improves the performance. Our results achieve state-of-the-art performance using a convolutional-recurrent architecture, whose parameters are refined with a discriminator. 

This paper is structured as follows. Section \ref{sec:related_work} reviews the state-of-the-art models for visual saliency prediction and recent advances on conditional adversarial networks. Section \ref{sec:architecture} presents PathGAN, our deep convolutional-recurrent neural network, as well as the discriminator network used during the adversarial training. Section \ref{sec:training} describes the training procedure and the loss functions used. Section \ref{sec:experiments} includes the experiments and results of the described techniques. Finally, Section \ref{sec:conclusions} discusses the main conclusions and future work. Our results can be reproduced with the source code and trained models available at \url{https://github.com/imatge-upc/pathgan}.
\section{Related Work}
\label{sec:related_work}

\subsection{Visual Saliency Prediction}

\textbf{Saliency maps.} Saliency prediction has received interest by the research community for many years. Thus seminal works by Itti et al. \cite{Itti1998PAMI} proposed considering low-level features at multiple scales and combining them to form a two-dimensional saliency map. Harel et al. \cite{harel2006nips}, also starting from low-level feature maps, introduced a graph-based saliency model that defines Markov chains over various image maps, and treat the equilibrium distribution over map locations as activation and saliency values.  Judd et al. in \cite{judd2009iccv} presented a bottom-up, top-down model of saliency based not only on low but mid and high-level image features. Borji \cite{borji2012cvpr} combined low-level features saliency maps of previous best bottom-up models with top-down cognitive visual features and learned a direct mapping from those features to eye fixations.

As in many other fields in computer vision, a number of deep learning solutions have very recently been proposed that significantly improve the performance.
For example, the Ensemble of Deep Networks (eDN) \cite{vig2014large} represented an early architecture that automatically learns the representations for saliency prediction, blending feature maps from different layers. Their network might be consider a shallow network given the number of layers. In \cite{Pan_2016_CVPR} shallow and deeper networks were compared. DCNN have shown better results even when pre-trained with datasets build for other purposes. DeepGaze \cite{kummerer2015deep} provided a deeper network using the well-know AlexNet \cite{krizhevsky2012imagenet}, with pre-trained weights on Imagenet \cite{deng2009imagenet} and with a readout network on top whose inputs consisted of some layer outputs of AlexNet. The output of the network is blurred, center biased and converted to a probability distribution using a softmax. Huang et al.~\cite{huang2015salicon}, in the so call SALICON net, obtained better results by using VGG rather than AlexNet or GoogleNet \cite{Szegedy2015}. In their proposal they considered two networks with fine and coarse inputs, whose feature maps outputs are concatenated.

Li et al. \cite{Li_2015_CVPR} proposed a multi resolution convolutional neural network that is trained from image regions centered on fixation and non-fixation locations over multiple  resolutions. Diverse top-down visual features can be learned in higher layers and  bottom-up visual saliency can also be inferred by combining information over multiple resolutions. 
Cornia et al. \cite{mlnet2016} proposed an architecture that combines features extracted at different levels of a DCNN. They introduced a loss function inspired by three objectives: to measure similarity with the ground truth, to keep invariance of predictive maps to their maximum and to give importance to pixels with high ground truth fixation probability. In fact choosing an appropriate loss function has become an issue that can lead to improved results. Thus, another interesting contribution of Huang et al.~\cite{huang2015salicon} lies on minimizing loss functions based on metrics that are differentiable, such as NSS, CC, SIM and KL divergence to train the network (see \cite{riche2013iccv} and  \cite{kummerer2054IG} for the definition of these metrics. A thorough comparison of metrics can be found in \cite{Bylinskii2016metrics}). 
In Huang's work \cite{huang2015salicon} KL divergence gave the best results. Jetley et al. \cite{jetley2016end} also tested loss functions based on probability distances, being
the Bhattacharyya distance the one that provided the best results. 



\textbf{Scanpaths.} The literature on the related task of scanpath prediction is much smaller, but has received recent attention caused by the rise of VR and AR technologies \cite{rai2017saliency}. In \cite{cerf2008predicting}, Cerf et al. concluded that human observers -- when not instructed to look for anything in particular -- tend to fixate on a human face within the first two fixations with a probability over 80\%. Moreover, the consistency of scanpaths increases when faces are present. Hu et al. \cite{hu2017deep} introduced a model that predicts relevant areas of a 360-degree video and decides in which direction a human observer should look for each frame. Some authors have also focused on omni-directional images \cite{rai2017saliency,yucheng2018prediction,ling2018saliency}.

SalTiNet \cite{assens2017saltinet} proposed a deep learning approach that proposes a novel three-dimensional representation of saliency maps: the \textit{saliency volumes}. This data structure captured the temporal location of the fixation across an additional temporal axis added to the classic saliency maps. The final scanpath are generated by sampling fixation points from this saliency volumes and finally introducing a post-filtering stage.
PathGAN also uses a deep neural model, but provides a fully end-to-end solution where the model directly generates a scanpath, with no need of any sampling nor post-processing.


\subsection{Generative Adversarial Networks}

The generation of a sequence of fixation points over an image with a Recurrent Neural Network (RNN) had been previously attempted in \cite{assens2017saltinet}. The authors trained a RNN to minimize the $L^2$ loss between predicted and ground truth scanpaths, but the resulting model tended to predict output fixations always in the center, as this is the best option on average for that loss function. Similar problems have been observed in other image prediction problems (e.g. \textit{pix2pix}), where blurred images where output as a result \cite{pathak2016context,mathieu2015deep,zhao2016energy}. 

The generation of diverse and realistic new data samples has received a lot of interest thanks to the work of Ian Goodfellow et al. on Generative Adversarial Networks (GANs) \cite{goodfellow2014generative}. 
In this framework, two models are trained iteratively. First, the generative model \textit{G} tries to capture the data distribution. Second, the discriminator model \textit{D} estimates the probability that a given sample is synthesized or real. During training, \textit{G} tries to maximize the probability of fooling \textit{D}. This process can also be seen as if GANs learn a loss function to tell if a sample is real or fake. Generated samples that are not realistic (e.g. blurry images, or scanpaths with all the fixations in the center) will not be tolerated.

A popular variation of GANs are the Conditional Adversarial Networks (cGANs) \cite{radford2015unsupervised}, where \textit{G} does not output a sample purely from a noise vector, but it is also conditioned on a given input vector. In this setting, \textit{D} needs to observe the conditioning vector to decide about the nature of the sample to be classified into synthesized or real. 
There have been multiple variations around the cGAN paradigm. Isola et al. \cite{isola2017image} proposed cGANs as a general purpose solution for image-to-image translation tasks using a \textit{U-Net}~\cite{ronneberger2015u} architecture for the generator, and a convolutional \textit{PatchGAN} \cite{li2016precomputed} architecture for the discriminator. Reed et al. bridge recent advances in the image and text fields and propose a GAN architecture that is capable of generating plausible images of birds and flowers from detailed text descriptions \cite{reed2016generative}. Mirza et al. conditioned GANs to discrete labels in order to generate MNIST digits conditioned on class labels \cite{mirza2014conditional}. Gauthier et al. generates faces with specific attributes by varying the conditional information provided to the network \cite{gauthier2014conditional}.

In our work, we adopt the cGAN paradigm to overcome the limitation reported in \cite{assens2017saltinet} when trying to use a RNN for visual scanpath prediction. This way, PathGAN proposes to train a RNN following an adversarial approach, in such a way that the resulting generator produces realistic and diverse scanpaths conditioned to the input image.

\section{Architecture}
\label{sec:architecture}

\begin{figure}[h]
\centering
\includegraphics[width=\textwidth]{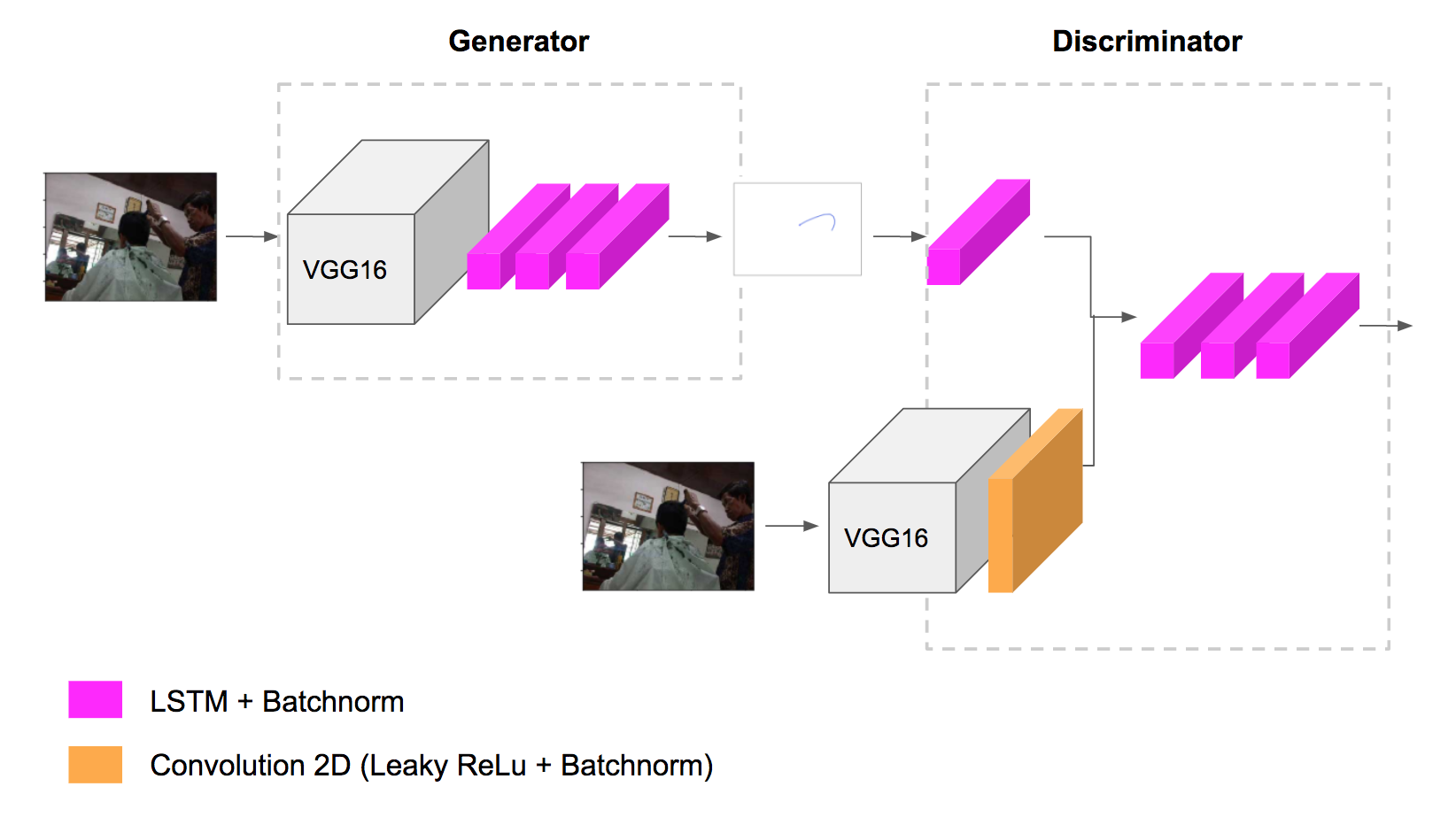}
\caption{Overall architecture of the proposed convolutional-recurrent model}
\label{fig:model}
\end{figure}

The overall architecture of PathGAN is depicted in Figure \ref{fig:model}. It is composed by two deep neural networks, the generator and the discriminator, whose combined efforts aim at predicting a realistic scanpath from a given image. 
The model is trained following the cGAN framework to allow the predictions to be conditioned to an input image, encoded by a pre-trained convolutional neural network. This section provides details about the structure of both networks and the considered loss functions.

\subsection{Objective}
\label{subsec:objective}

GANs are generative functions that learn a transformation from random noise vectors $z$ to output vectors $y$, $G:z \rightarrow y$ \cite{goodfellow2014generative}. Conditional GANs learn a transformation from a given input vector $x$ and random noise vector $z$, to $y$, $G: {x, z} \rightarrow y$. Therefore, the objective function of cGANs can be expressed as: 
\begin{equation}
L_\mathrm{cGAN} (G, D) =
\mathbb{E}_{x,y}[\log D(x, y)]+ \mathbb{E}_{x,z}[\log(1 - D(x, G(x, z))],
\label{eq:cgan}
\end{equation}
where the generator tries to minimize the loss, while the discriminator tries to maximize it. 

\textbf{Multi-objective loss functions.} Previous works has found useful to mix the GAN's loss function with another traditional loss such as the Euclidean distance \cite{pathak2016context}. In this case, the task of the discriminator remains unchanged, but the generator is forced to output samples that are close to the ground truth (in terms of $L^2$ distance). We found that this setting improved stability and convergence rate of the adversarial training. As it will be explained in the next section, each prediction of our model contains four dimensions. The $L^2$ distance is computed using all four dimensions. We called this parameter \textit{content loss}, and it is defined as:
\begin{equation}
L_{L^2}(G) = \mathbb{E}_{x,y,z}[\|y - G(x, z)\|^2].
\end{equation}
The final formulation of the loss function for the generator during adversarial training is: 

\begin{equation}
L = L_\mathrm{cGAN} (G, D) + \alpha · L_{L^2}(G).
\end{equation}

In Equation \ref{eq:cgan}, $(1 - D(G(x, z))$ represents the probability of the generator fooling the discriminator. Thus, we expect the loss to decrease as the chances of fooling the discriminator increase. In our experiments we used the hyperparameter $\alpha = 0.05$. It is also important to note that $z$ plays an important role making the output of the generator non-deterministic \cite{wang2016generative}. During the training of the discriminator the content loss is not used.

\subsection{Generator}
\label{subsec:generator}

The generator reads images as input and outputs a variable length sequence of predicted fixation points. In addition to the coordinates of the fixation points, our model has an end-of-sequence (EOS) neuron to encode the scanpath variable length behavior. This neuron has values between $[0, 1]$ and represents the probability of having reached the end of the sequence. Thus, each prediction of our model contains a fixation point (composed by a spatial coordinate and a timestamp) and an EOS parameter  $[x, y, t, \mathrm{EOS}]$. At training time, we train on fixations of a scanpath until we reach the EOS, and at test time we predict scanpath fixations until we reach the EOS. 

We propose a convolutional-recurrent architecture that learns its filter parameters to predict scanpaths. Figure \ref{fig:model} illustrates the architecture of the model, composed of 49 million free parameters. The generator is composed of two parts. First, high-level image features are extracted using a convolutional neural network for image recognition named VGG16 \cite{simonyan2014very} pre-trained on the ImageNet dataset \cite{deng2009imagenet}. Then, resampling of the VGG16 activations is performed with an Average Pooling layer to a fixed size representation. This allows the usage of this model with different image sizes and different types of datasets. Finally, a recurrent module composed of 3 fully connected LSTMs with tanh activation and 1,000 hidden units is used to generate a variable length scanpath. Batch normalization layers are placed after each recurrent layer to improve convergence and accelerate learning.

\subsection{Discriminator}
\label{subsec:discriminator}

Figure \ref{fig:model} also shows the architecture and layer configuration of the discriminator. This network predicts if a given scanpath is synthesized or not, and this decision is conditioned to the associated image. 

It is clear that knowledge of the image that a scanpath corresponds to is essential to evaluate quality. Moreover, previous work has shown that conditioning the discriminator function to the input significantly increases the performance, sometimes preventing the generation from collapse \cite{isola2017image}. In our architecture, the discriminator has two input branches; a branch where a scanpath is read, and a branch where the image is read. This allows discriminating whether a scanpath is realistic for a given image. The features of the two branches are concatenated. 

Briefly, the discriminator function is based on a recurrent architecture where the scanpath fixations are read sequentially. The network is composed of a VGG16 module that extracts image features, and three recurrent layers interspersed with batch normalization layers. The recurrent layers contain 1000 hidden units and a tanh activation. Similarly to the generator, the VGG16 activations are resampled with an Average Pooling layer to a fixed size representation. The recurrent layers all use $tanh$ activations, with the exception of the final layer, which makes use of a sigmoid activation.
\section{Training}
\label{sec:training}
The weights of the model have been learned with an objective function that combines an adversarial loss and a content loss \cite{isola2017image}. The content loss follows a simple approach in which the generated and ground truth fixation points are compared using the $L^2$ norm (or mean square error). The adversarial loss depends on the probability of the generator fooling the discriminator.  

We trained the PathGAN architecture on two datasets. First, the network was trained using the iSUN dataset, which contains 6,000 training images. Then, the filter weights were fine-tuned on omni directional images using the Salient360 dataset, which has 40 training images. For validation purposes, we split the training data into 80\% for training and the rest for validation. Notice that for each gradient update a single scanpath is used. 

The spatial positions of the fixations were normalized to $[0, 1]$. Moreover, when training on the Salient360 dataset, input images were downsampled to fit the dimensions of $300 \times 600$ prior to training. We also subtracted the mean pixel value of the training set from the image's pixels to zero center them.

The architecture was trained using the $RMSprop$ optimizer with the following settings: $lr = 10^{-4}$, $\rho=0.9$, $\epsilon = 10^{-8}$ and without decay. 

\begin{figure}[h]
\centering
\includegraphics[width=0.9\textwidth]{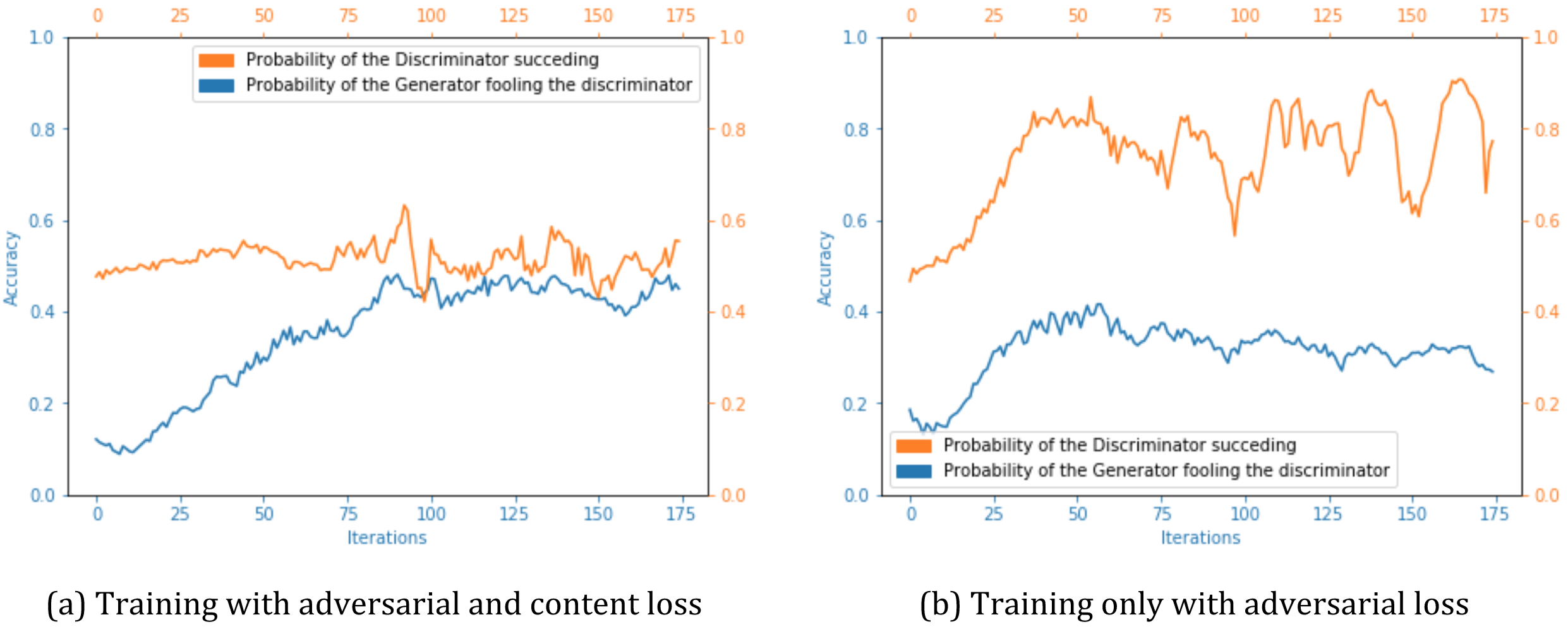}
\caption{iSUN validation set accuracies for training with GAN+MSE vs GAN on varying number of epochs.}
\label{fig:training}
\end{figure}

Our network took approximately 72 hours to train on six NVIDIA Tesla K80 GPU using the Keras framework with Tensorflow backend. At test time it generates approximately 4 scanpaths per second. Figure \ref{fig:training} shows the evolution of the validation set accuracy during the adversarial training.

Our networks train on a minibatch size of $m = 100$, and after trying various combinations, we settled on the generator doing 8 gradient updates, while the discriminator does 16 for each iteration. At train time, the generator is first bootstrapped by training only on the content loss for a duration of 5 epoch. Then, the adversarial training begins.

This architecture was designed considering the amount of training data available, and multiple strategies were introduced to prevent overfitting. In the first place, the convolutional modules initialized from the VGG16 model were not fine-tunned, decreasing the number of training parameters. Second, the input images were resized to a smaller dimension, and dropout noise was introduced at training time. We also used dropout noise ($p = 0.1$) on the recurrent layers.

\section{Experiments}
\label{sec:experiments}
PathGAN was assessed and compared from different perspectives.
First, we evaluated the performance on traditional images using the iSUN dataset. Second, we show quantitative performance results on omni directional images using the Salient360 dataset.

\subsection{Datasets}
The network was initially trained on the iSUN dataset \cite{xu2015turkergaze} that contains 6,000 training images, and its performance is benchmarked in Section \ref{subsec:results}. Then, the network was fine-tuned to predict scanpaths on omni directional images using the Salient360 dataset, which contains 60 training images with data obtained from head and eye movements from the human observers.

It is worth noticing that our use of omni directional images in this network implies an important simplification. We assume that omni-directional images are similar to traditional flat images, just with a bigger size. This presents advantages like being able to reuse the same architecture, and easily fine-tune it, and this strategy has been previously successful \cite{assens2017saltinet}. Nevertheless, it neglects the characteristic of omni directional images where points that are close to opposite corners are spatially close. 

\subsection{Metrics}
\label{subsec:metrics}

The similarity metric used in the experiments is the Jarodzka algorithm \cite{jarodzka2010vector}. This metric presents different advantages over other common metrics like the Levenshtein distance or correlating attention maps. In the first place, it preserves the overall shape, direction and amplitude of the saccades, the position and duration of the fixations. Second, it provides more detailed information on the type of similarity between two vectors. This metric has been recently used in the Salient360, scanpath prediction challenge at ICME 2017 \cite{salient360}. The implementation of the metric for omni directional images was released by the University of Nantes \cite{Gutierrez2018Toolbox}. This code was adapted to compute the Jarodzka metric for conventional images on the iSUN dataset.

The ground truth and predicted scanpaths are then matched 1-to-1 using the Hungarian algorithm to obtain the obtain the minimum cost. The presented results compare the similarity of 40 generated scanpaths with scanpaths in the ground truth. 

\subsection{Results}
\label{subsec:results}

\subsubsection{Comparison with state-of-the-art}

PathGAN is compared using the iSUN and Salient360! datasets. Table \ref{tab:icme} compares the performance on omni directional images using the Jarodzka metric, against other solutions presented at the Salient360! Challenge \cite{salient360}, which took place at the IEEE ICME 2017 conference in Hong Kong. The results of the participants were calculated by the organization, on a test set whose ground truth was not public at the time. Although at the time of writing this test set is public, our model has only been trained on the training set. These results indicate the superior performance of PathGAN with respect to the participants.

\begin{figure}[h!]

\subfloat[Mean performance on iSUN with the Jarodzka metric]{

\begin{tabular}{llc}
\hline
id &			& Jarodzka$\downarrow$ \\
\hline
a& Random positions and number of fixations & 0.71\\
b& Random positions and GT number of fixations & 0.45\\
c&Sampling ground truth saliency maps & 0.31\\
d&Interchanging scanpaths across images & 0.23 \\
\hline
e&SalTiNet & 0.69\\
f&PathGAN without content loss & 0.42 \\
g&SalTiNet (fine-tuned on iSUN) & 0.40\\
\textbf{h}&\textbf{PathGAN} & \textbf{0.13}\\
\hline

\end{tabular}
}
\par
\subfloat[Distribution of results obtained for each model]{\includegraphics[clip,width=1.0\columnwidth]{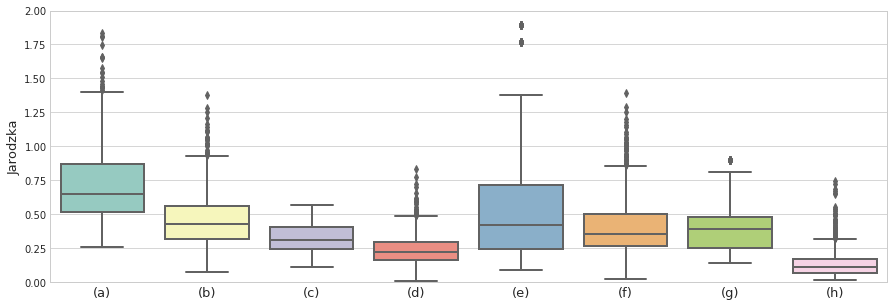}}
\caption{Comparison on iSUN between the state-of-the-art and baselines. The distribution of results and the mean performance are depicted. Lower values are better.}
\label{tab:isun}
\end{figure}

\begin{figure}[h!]
\centering
\includegraphics[width=0.5\textwidth]{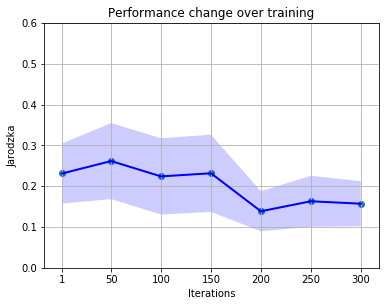}
\caption{iSUN validation set Jarodzka evaluation on varying number of mini-batches.}
\label{fig:jar_evo}
\end{figure}

Figure \ref{tab:isun} compares the performance of PathGAN with different baselines and another state-of-the-art model on the iSUN dataset.
To accurately test the performance of the best scanpath prediction model of the Salient360! Challenge 2017 on the iSUN dataset, we fine-tuned it. Figure \ref{fig:jar_evo} illustrates how the Jarodzka performance of PathGAN evolves during training. 

\vspace{-10mm}

\clearpage

\setlength{\tabcolsep}{10pt}
\begin{table}[h!]
\begin{center}
\begin{tabular}{lccccc}
\hline
 & Wuhan University  &  SJTU & SaltiNet & \textbf{PathGAN} \\
\hline
Jarodzka $\downarrow$ &  5.9517 & 4.6565 & 2.8697 & \textbf{0.74} \\
\hline
\end{tabular}
\end{center}
\caption{Comparison with the best submissions to the
ICME 2017 Salient360! Lower values are better. }
\label{tab:icme}
\end{table}

\subsubsection{Content-loss gain}
The performance gain that comes with the use of a content-loss based on MSE was analyzed from different perspectives. Figure \ref{fig:training} shows that the \textit{content loss} (mentioned in Section \ref{subsec:objective}) significantly improves convergence. In our experiments, we have not been able to achieve convergence without using the MSE loss. Table \ref{tab:isun} illustrates that these improvements are also reflected in the Jarodzka metric.

\subsubsection{Qualitative results}

\begin{figure}[h!]
\centering
\includegraphics[width=0.8\textwidth]{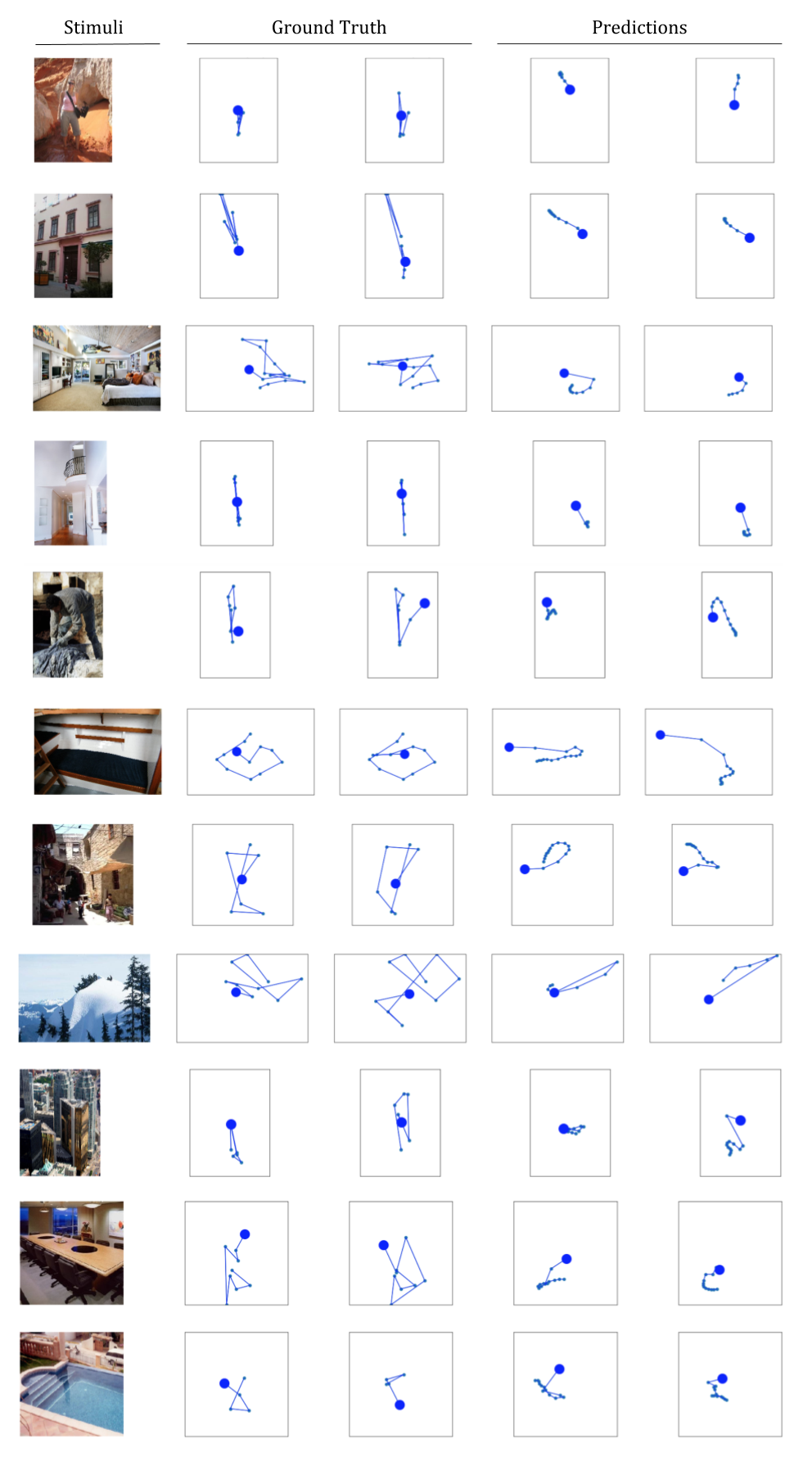}
\caption{Examples of predictions and ground truth on the iSUN dataset.}
\label{fig:ex_isun}
\end{figure}

\begin{figure}[h!]
\centering
\includegraphics[width=0.9\textwidth]{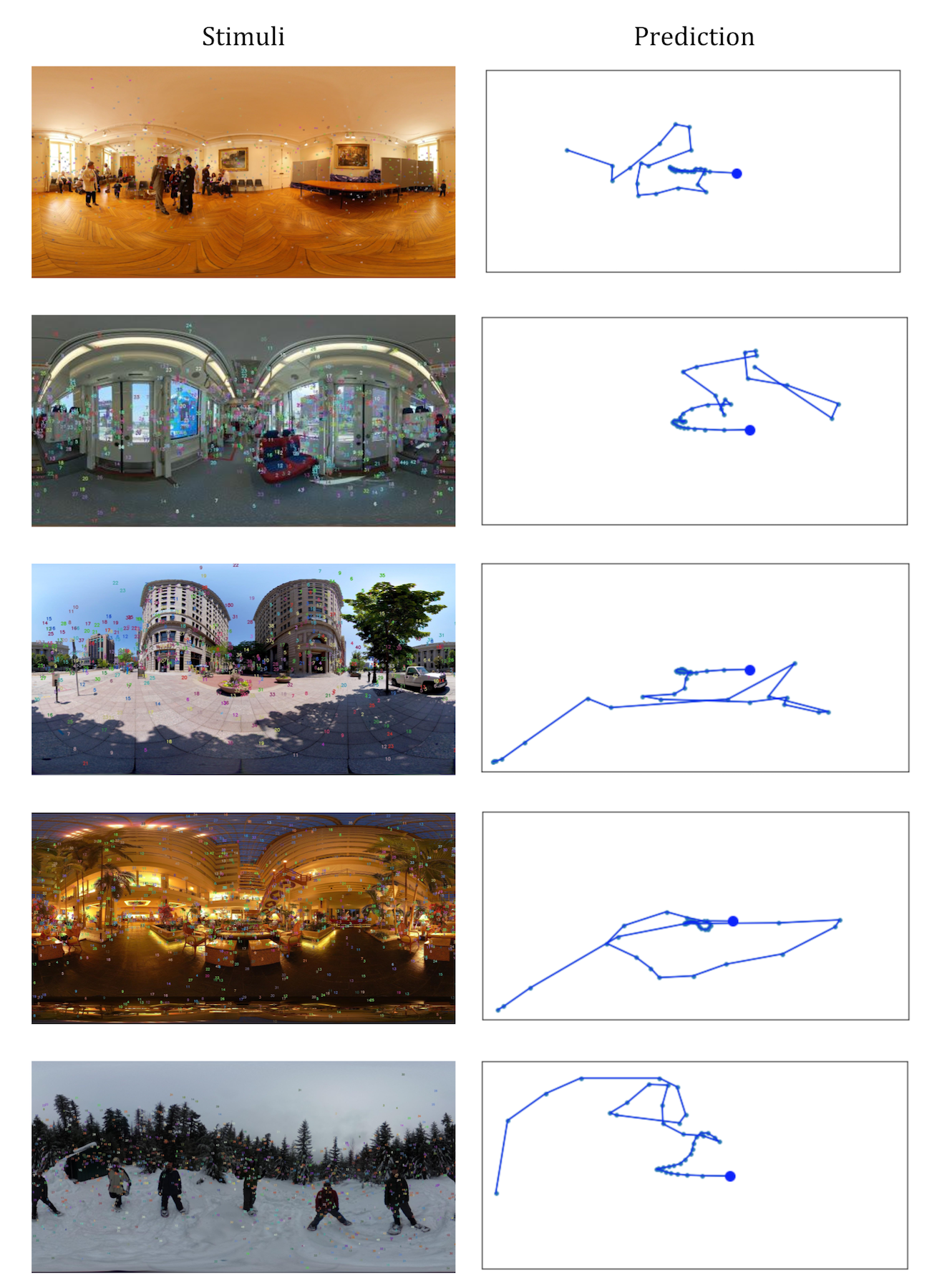}
\caption{Examples of predictions on the Salient360! dataset. The stimuli has the ground truth annotated.}
\label{fig:ex_360}
\end{figure}

Our model's performance has also been explored from a qualitative perspective by observing the generated scanpaths on the iSUN dataset  and on the Salient360! dataset (Figures \ref{fig:ex_isun} and \ref{fig:ex_360}). Notice the diversity of results given the generative nature of the model, based on the drop out ratio in the LSTM.

\begin{figure}[h]
\centering
\includegraphics[width=0.7\textwidth]{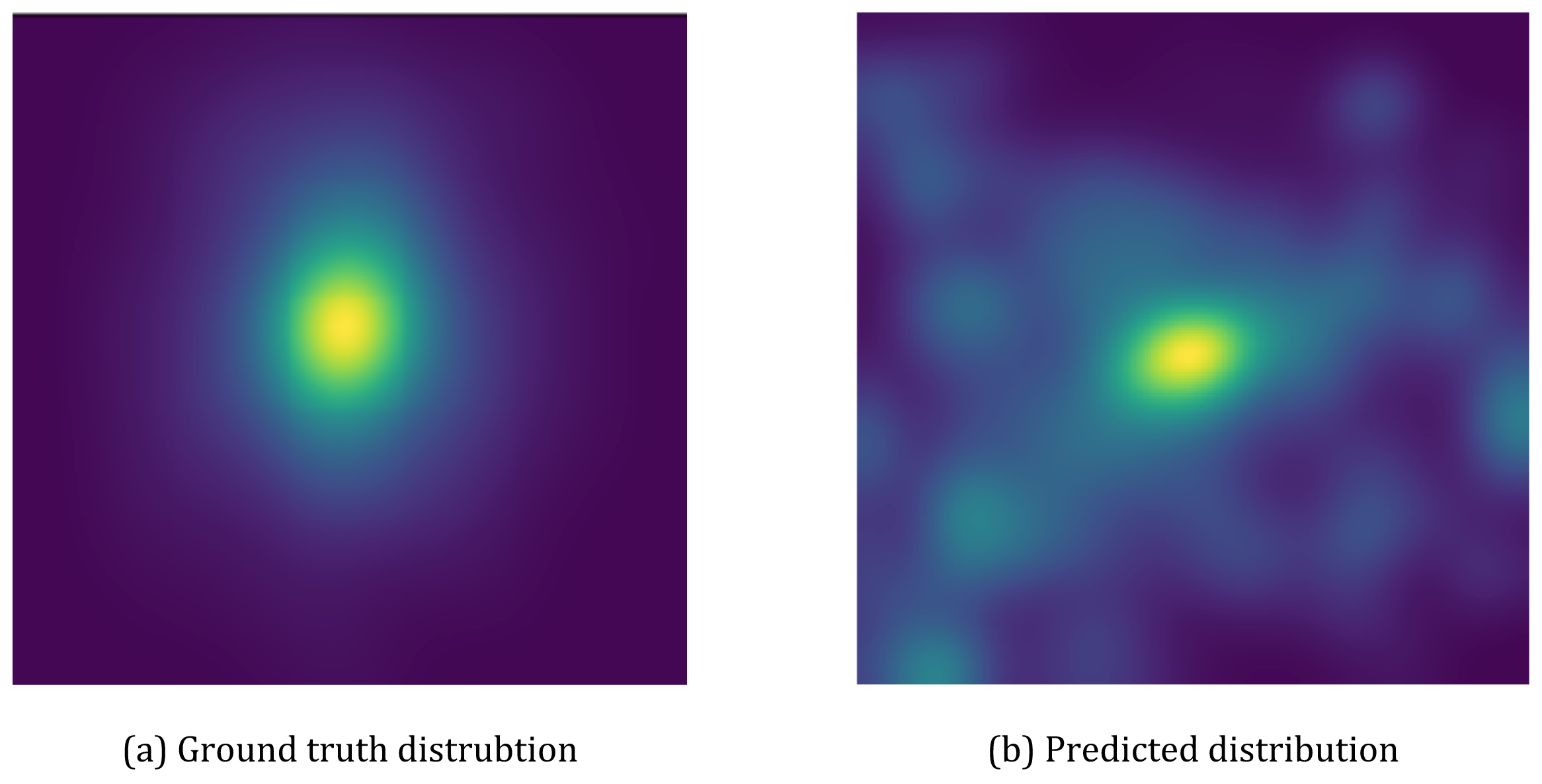}
\caption{Comparison of generated and ground truth spatial distribution of  fixations}
\label{fig:agg}
\end{figure}

Another way of assessing the behaviour of our model is by comparing the distributions of generated and ground truth fixations. Figure \ref{fig:agg} compares the distribution of spatial locations where the model fixates on the iSUN's validation dataset. We observe that the model correctly finds a center-bias.

\clearpage
\section{Conclusions}
\label{sec:conclusions}
Most of the work that has been done in the field of saliency estimation focuses on aggregating fixations from multiple observers and the prediction of saliency maps. Thus, it does not pay any attention to the temporal dimension of saliency estimation. This paper addressed a task that is closer to what a human does when observing an image: \textit{scan path prediction}. This task presents several challenges, such as the complicated distribution of the data, and we address them accordingly. 

We presented PathGAN, an end-to-end model capable of predicting scanpaths on ordinary and omni-directional images using the framework of conditional adversarial networks. Our experiments show that this architecture achieves state-of-the-art results on both scenarios. Moreover, this model has the following desirable characteristics: 1) the probability of a fixation is conditioned to previous fixations; and 2) the length of the scanpath, the duration of each fixation, and the spatial position of the fixations are treated as conditioned random variables. 

Finally, we want to note that the use of this model with omni-directional images assumes the simplification that an omni-directional image is similar to a traditional image, but with a larger size. While this presents advantages, it also has a drawback: it neglects the characteristic of omni-directional images where points that are close to opposite corners are spatially close. 

Future work could aim to solve the issue mentioned above, or could try to include top-down task specific information during training. 
Our results can be reproduced with the source code and trained models available at \url{https://github.com/imatge-upc/pathgan}.

\clearpage

\bibliographystyle{splncs}
\bibliography{egbib}

\end{document}